\documentclass{Interspeech2024}

\usepackage{cite}
\usepackage{comment}
\usepackage{threeparttable}
\usepackage{multirow}
\newcommand{\argmax}{\mathop{\rm arg~max}\limits}

\interspeechcameraready 

\title{Factor-Conditioned Speaking-Style Captioning}

\name{Atsushi}{Ando}
\name{Takafumi}{Moriya}
\name{Shota}{Horiguchi}
\name{Ryo}{Masumura}

\address{
  NTT Corporation, Japan
  }
\email{atsushi.ando@ntt.com}

\keywords{speaking-style captioning, large language models, pre-trained speech encoder, factor-conditioned captioning}

\begin{document}

\maketitle

\begin{abstract}
This paper presents a novel speaking-style captioning method that generates diverse descriptions while accurately predicting speaking-style information. Conventional learning criteria directly use original captions that contain not only speaking-style factor terms but also syntax words, which disturbs learning speaking-style information. To solve this problem, we introduce factor-conditioned captioning (FCC), which first outputs a phrase representing speaking-style factors (e.g., gender, pitch, etc.), and then generates a caption to ensure the model explicitly learns speaking-style factors. We also propose greedy-then-sampling (GtS) decoding, which first predicts speaking-style factors deterministically to guarantee semantic accuracy, and then generates a caption based on factor-conditioned sampling to ensure diversity. Experiments show that FCC outperforms the original caption-based training, and with GtS, it generates more diverse captions while keeping style prediction performance.
\end{abstract}

\section{Introduction}
Understanding the speaking styles of speakers, including the speaker's attributes and prosodic information, is important in speech applications.
These applications include flexible control in speech synthesis~\cite{PromptTTS, PromptTTSpp, TextrolSpeech} or voice conversion~\cite{PromptVC}, understanding and controlling speaking behavior for human-like spoken dialogue systems~\cite{Estimating_User_Communication, Advancing_Large_Language}, and supporting other speech processing technologies, such as using auxiliary labels for recognizing non-linguistic or para-linguistic information~\cite{Selective_Multi-Task_Learning, Multi-Lingual_Multi-Task}.

Traditional studies formulate speaking-style recognition as audio classification into pre-defined speaking-style classes such as style categories, e.g., spontaneous or infant-directed~\cite{Characteristics_of_Speaking, Towards_Automatic_Classification}.
However, their applicability is limited due to the difficulty of predefining diverse speaking styles in real environments.
In response, recent studies have tackled speaking-style captioning, which predicts speaking styles in a free descriptive format~\cite{StyleCap, LTU-AS, Qwen-Audio}.
This eliminates the need to predefine speaking-style classes, since detailed speaking-style information, such as intensity and detailed expressions, can be used instead.
Conventional style captioning models are composed of a speech encoder, bridge network, and text decoder.
The speech encoder first extracts speech features from the input audio; the bridge network then converts these features into representations suitable for input to the text decoder, which generates a speaking-style caption from the representations.
Pre-trained autoregressive large language models~(LLMs) are used as the text decoder to generate natural and diverse captions.

One of the challenges of speaking-style captioning is to generate captions with accurate speaking-style factors, such as information of gender, pitch, power, and speaking speed.
However, using the conventional training method based on cross-entropy with ground-truth captions can make it difficult to ensure that the captioning model acquires speaking-style knowledge.
The ground-truth captions contain not only speaking-style factor terms but also syntax words, which disturbs to focus solely on learning the correct speaking styles.
Furthermore, even in the inference step, the conventional models suffer from errors in the prediction of speaking style factors if sampling-based decoding~\cite{The_Curios_Case} is used.
While such decoding is widely used to ensure caption diversity, it can lead to the occurrence of less probable words, potentially generating words that represent incorrect style factors.

This paper presents a simple but effective model training method named \textit{factor-conditioned captioning~(FCC)} that can predict speaking-style factors accurately.
The key idea of FCC is to make the captioning model explicitly predict only style factors as intermediate reasoning steps before generating style captions, like the Chain-of-Thought Prompting approach~\cite{Chain_of_Thought}.
The ground truth of FCC contains a factor phrase, the phrase representing speaking style factors, whose format is fixed, e.g., \textit{``male, low pitch, high volume, normal speed''}, in the first part, and the original caption in the second part.
The factor phrase can be generated by predicting the speaking-style factors from the original caption using GPT~\cite{GPT-4}, or by the ground-truth factors if available.
Our FCC-based training proposal forces the text decoder to consider style factors in generating captions since the decoder predicts the next word based on the preceding text.
One of the advantages of FCC is that it is applicable simply by changing the output sentence in the training step, regardless of the model structure or loss function.

For decoding, we introduce \textit{greedy-then-sampling~(GtS)} to generate diverse captions while considering the style factors.
GtS predicts the factor phrase by using a maximum likelihood criterion, followed by random sampling for the caption part.
This enables the generation of diverse captions in terms of expressions and syntax while the conditioning factors are determined by the hypothesis with the highest probability, which prevents the generalization of incorrect style factors.

Experiments on the PromptTTS dataset~\cite{PromptTTS} demonstrate that FCC significantly improves the performance of both the speaking-style captioning and individual speaking-style factor classification derived from captions compared to the conventional method.
Furthermore, FCC with GtS decoding yields more diverse captions while preserving accuracy of speaking style factor recognition.
Note that we conduct comprehensive comparisons of the components of the captioning model to make a strong baseline for speaking style captioning, and thus validate the superior effectiveness of the proposed methods.

\section{Related work}
Speaking-style captioning is often included as a task within speech understanding, enabling a single model to address multiple speech-related tasks.
Several prior studies proposed speech understanding models\cite{LTU-AS, Qwen-Audio, SpeechPrompt, Pengi, SLM, VIOLA, AudioPaLM, SECap}.
Though the previous studies employed various types of modules in the individual components, few have reported comparisons of the components.
To rectify this omission, this study conducts comprehensive evaluations of multiple methods for each of the three components in speaking-style captioning to investigate the optimal model structure and how each component affects the final result. 
Note that the proposed methods, including FCC and GtS, show significant improvements in the best model.

\section{Proposed speaking-style captioning}
Let $\bm{S}, \bm{I}$ be an input speech and an input instruction sentence to specify the task, with $\bm{O}$ is being corresponding output sentence.
Speaking-style captioning is formulated as the problem of predicting output sentence $\bm{O}$ from $\bm{S}$ and $\bm{I}$, 
\begin{equation}
  \bm{\hat{O}} = \argmax_{\bm{O}} P \left(\bm{O} \mid \bm{S}, \bm{I}; \bm{\Theta} \right),
\end{equation}
where $\bm{\Theta}$ is the set of captioning model parameters.
The instruction sentence is fixed to learn a single task, e.g., speaking style captioning.

\subsection{Captioning model}
\begin{figure}[t]
  \centering
  \includegraphics[width=\linewidth]{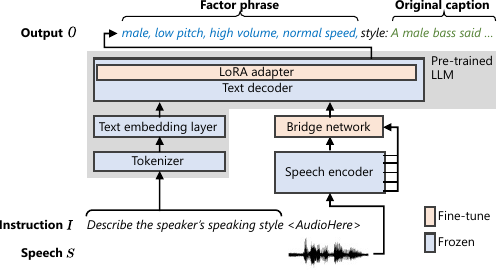}
  \caption{The outline of the proposal, FCC.}
  \label{fig:proposed_model}
\end{figure}

Similar to the conventional studies~\cite{StyleCap, SECap, Pengi, LTU-AS, SLM}, the captioning model consists of three blocks: speech encoder, bridge network, and text decoder, as shown in Figure~\ref{fig:proposed_model}.

The speech encoder extracts multiple sequences of speech representation vectors $\bm{Z} = [\bm{z}_1, \dots, \bm{z}_K]$ from the input speech,
\begin{equation}
  \bm{Z} = \mathsf{SpeechEnc}\left(\bm{S} ; \bm{\Theta}_s \right),
\end{equation}
where $\bm{z}_k \in \mathbb{R}^{V \times T}$ denotes the $T$-length sequence of $V$-dimensional vectors output from the $k$-th layer of the speech encoder, which consists of $K$ layers.
$\bm{\Theta}_s$ denotes the set of parameters of the speech encoder that are pre-trained and frozen during the training of the captioning model.
The use of multiple intermediate layers enables the model to extract different types of speech information~\cite{WavLM, HuBERT}.

The bridge network converts the speech representations into speech embeddings suitable for input to the text encoder:
\begin{align}
  \bm{e} = \mathsf{Bridge}\left(\bm{Z} ; \bm{\Theta}_b \right),
\end{align}
where $\bm{e} \in \mathbb{R}^{D \times N}$ is the $D$-length $N$-dimensional vector.
$\bm{\Theta}_b$ represents the set of bridge network parameters. 
As the bridge network, we use a time and layer-wise Transformer~(TLTR)~\cite{Whisper-AT} so as to combine multi-level information from different layers of the speech encoders into a $V$-dimensional embedding of length 1, i.e., $N$=1.
Finally, the embedding is projected into the same dimensions as the input of the text decoder by a fully connected layer.

The text decoder predicts the posterior probabilities of the output sentence from the speech embeddings $\bm{e}$ and embedded input instruction sentence $\bm{v}$ as
\begin{align}
    \bm{v} &= \mathsf{TextEmb}\left(\mathsf{Tokenizer}\left(\bm{I}\right); \bm{\Theta}_t \right), \\ 
    P\left(\bm{O}\right) &= \mathsf{TextDec}\left(\bm{v}, \bm{e} ; \bm{\Theta}_d \right),
\end{align}
where $\bm{v} \in \mathbb{R}^{D \times M}$ and $M$ is the embedding sequence length.
$\bm{\Theta}_t, \bm{\Theta}_d$ represent the set of parameters for the text embedding layer and the text decoder, respectively. 
The proposed method employs an autoregressive text decoder and the corresponding text tokenizer derived from a pre-trained LLM such as LLaMA-2~\cite{LLaMA2}.
Furthermore, in the text decoder, an adapter based on low-rank adaptation~(LoRA)~\cite{LoRA} is inserted into the original decoder to adapt the output sentences of the training data. 
That is, the set of the text decoder parameters composed of the parameter sets of the adapter $\bm{\Theta}_{d_a}$ and the original LLM decoder $\bm{\Theta}_{d_o}$, where $\bm{\Theta}_d = {\bm{\Theta}_{d_o}\cup\bm{\Theta}_{d_a}}$.
The parameters of $\bm{\Theta}_t$ and $\bm{\Theta}_{d_o}$ are frozen during training, while $\bm{\Theta}_{d_a}$ are updated.

\subsection{Training by factor-conditioned captioning}
\label{sec:FCC}
\begin{figure}[t]
  \centering
  \includegraphics[width=\linewidth]{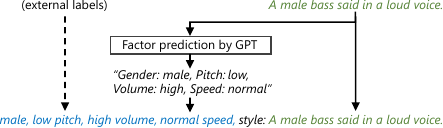}
  \caption{Generation of the ground-truth output in FCC.}
  \label{fig:gen_fcc}
\end{figure}

To help the captioning model learn speaking-style factors explicitly in addition to syntactic information, FCC uses the extended version of the original captions as ground truth.
Each extended caption consists of two parts: factor phrase, i.e., comma-separated terms, each of which represents a speaking-style factor, as the first part, and the original caption as the second part.
The use of this output enables the captioning model to predict the speaking-style information derived from speech first and generate the caption with explicit consideration of speaking style factors since the text decoder is autoregressive.

Readers may find it difficult to pre-define such factors, but this study avoids that difficulty by automatically generating them from the original captions alone.
As shown in Figure~\ref{fig:gen_fcc}, speaking-style factors are predicted by GPT~\cite{GPT-4} with a specific prompt\footnote{``\textit{Please select the information contained in the input sentence. The options are as follows:\texttt{<br>}Gender: male or female\texttt{<br>} Pitch: low, normal, high (normal if unspecified)\texttt{<br>}Volume: low, normal, high (normal if unspecified)\texttt{<br>}Speaking Speed: slow, normal, fast (normal if unspecified)\texttt{<br>}\texttt{<br>}Input: \{caption\}.}'', where \texttt{<br>} represents line break.}, and so a template is used to create the factor phrase.
The factors unspecified in the caption are regarded as \textit{normal}.
The template of the factor phrase is fixed during training, which makes the captioning model focus on predicting the factor information.
Of course, if ground-truth style factor labels are available, they can be used instead of predictions by GPT.

\subsection{Greedy-then-sampling decoding}
The common sampling strategies are not suitable for FCC because there is a risk of outputting style factors with low probability, which will yield caption errors.
Thus, we also propose GtS, which is a decoding strategy for FCC in generating captions with diverse expressions.
In GtS, decoding begins with a greedy search criterion until a delimiter token that connects the factors and caption (`\textit{style:}' in Figure~\ref{fig:proposed_model}) is generated.
Then, it switches to a sampling criterion~\cite{The_Curios_Case} to generate the main speaking style caption.
The greedy search ensures the correctness of the style factors by simply using those of the highest posterior probabilities, while the latter sampling allows the generation of diverse captions constrained on the predicted style factors.

\section{Experiments}

\subsection{Setup}
The PromptSpeech dataset provided by PromptTTS~\cite{PromptTTS} was used for our evaluations.
The audio is from the English Audiobook corpus, LibriTTS~\cite{LibriTTS}, and the labels include annotated speaking-style captions, as well as discrete speaking-style factors for gender~(male/female), pitch~(high/normal/low), volume~(high/normal/low), and speaking speed~(fast/normal/slow).
The dataset was split into the official speaker-open subsets as in the conventional work of~\cite{StyleCap}.
The training set consists of 24,953 utterances by 1,113 speakers, the development set of 857 utterances by 40 speakers, and the test set of 778 utterances by 38 speakers.
The distributions of each class of the style factors were roughly equal.
Note that from the total of 26,588 utterances, there were 154 unique captions.

The speech encoder was Whisper large-v3~\cite{Whisper}~(denote as Whisper-L), consisting of 32 intermediate layers with 1,280 units each; the input was fixed to 30 seconds, i.e., 1,500 frames.
As TLTR in the bridge network~(TLTR-utt), initial pooling of intermediate layer outputs was performed every 20 frames, and both the time and layer-wise Transformers were single-layer Transformers with 512 hidden units.
The text decoder was LLaMA-2 7B-chat~\cite{LLaMA2}, and the LoRA was the attention-based method~\cite{LoRA} with $r$=8 and $\alpha$=32~(7B-chat-LoRA).
The trainable/total parameters in each component were 0M/637M in the speech encoder, 90.7M/90.7M in the bridge network, and 4.2M/6.7B in the text decoder, respectively.
The conventional method was StyleCap~\cite{StyleCap} consisting of WavLM-base-plus~\cite{WavLM} speech encoder, weighted-sum + BLSTM + Q-Former~\cite{BLIP-2}-based bridge network, and LLaMA-2-7B decoder without LoRA.
The batch size was 16.
The optimizer was AdamW~\cite{AdamW} with learning rate of $1\times10^{-4}$.
The number of epochs was set to 5, with early stopping based on the development set.
Note that we did not pre-train the bridge network using the close-ended style classification tasks~\cite{LTU-AS}\footnote{Our preliminary experiments found there were no statistically significant improvements with bridge-network pre-training.}.
The baseline methods were the models trained by the original caption with greedy or sampling-based decoding.
Two methods of creating FCC were evaluated: the case when factor phrases were created using the predicted labels using GPT~(denoted as w/ predicted factors), and those created using discrete speaking-style factor labels provided in the dataset~(w/ golden factors).
The \textit{gpt-4-0613} model with the same prompt as described in Section~\ref{sec:FCC} was used for the former FCC.
We evaluated the proposed FCC with greedy, sampling, and GtS decoding.
In the inference step, the thresholds of the probability and the number were set to 0.9 and 40, respectively, for both sampling and GtS decoding.

The evaluation tasks included not only speaking-style captioning but also style factor classification.
The style factor classification involved using GPT to estimate factor classes for gender, pitch, volume, and speaking rate from captions, and then evaluating the accuracy against annotated discrete labels.
As the mapping from caption to four speaking style factors, the predicted caption was given to the \textit{gpt-4-0613} model with the same prompt as described in Section~\ref{sec:FCC}, then the output sentence was parsed and mapped to several non-target terms like `\textit{unspecified}' to `\textit{normal}' in post-processing.
Note that we created a codebook of GPT input/output pairs to ensure that the same factor prediction results for the same captions were returned in all the experiments to avoid the effect of sampling-based decoding on GPT.
The evaluation metrics for speaking-style captioning were BLEU-4~(B@4), ROUGE-L~(ROU), METEOR~(MET), BERTScore with \textit{distilbert-base-uncased}~\cite{distilbert}~(BS), and distinct-1 and 2~(dis-1/2), as a subset of the baseline study~\cite{StyleCap}.
The first four metrics were implemented by Hugging Face evaluate~\cite{evaluate}.

\subsection{Results}
\subsubsection{Main results}
\setlength{\tabcolsep}{1.5mm} 
\begin{table*}[t]
  \setlength{\tabcolsep}{1.31mm}
  \caption{
  Performance of speaking-style captioning and accuracy (\%) of style factor classification from captions using GPT.
  The baselines are our model, trained by original captions with greedy or sampling-based decoding.
  The best results are highlighted in \textbf{bold}.
  Those metrics calculated using the ground-truth captions are also shown at the bottom for reference.
  }
  \vspace{-1em}
  \label{tbl:main_results}
  \eightpt
  \centering
  \sisetup{detect-weight,mode=text}
  \renewrobustcmd{\bfseries}{\fontseries{b}\selectfont}
  \renewrobustcmd{\boldmath}{}
  \newrobustcmd{\B}{\bfseries}
  \begin{threeparttable}
  \begin{tabular}{@{}*{3}{l}*{6}{S[table-format=0.3]}*{5}{S[table-format=2.1]}@{}}
    \toprule
    &           &           & \multicolumn{6}{c}{Speaking-style captioning} & \multicolumn{5}{c}{Style factor classification} \\ \cmidrule(lr){4-9}\cmidrule(l){10-14}
    Model                          & Output    & Decoding    & {B@4$\uparrow$} & {ROU$\uparrow$} & {MET$\uparrow$} & {BS$\uparrow$} & {dis-1$\uparrow$} & {dis-2$\uparrow$} & {Gender}   & {Pitch}   & {Speed}   & {Volume}   & {Avg.}  \\  \midrule
    StyleCap~\cite{StyleCap}   & Caption   & Greedy    & 0.273         & 0.497         & 0.469         & 0.855        & 0.023 & 0.073 & 91.0  & 61.0  & 85.2  & 69.9  & 76.8  \\  \midrule
    Ours                      & Caption   & Greedy    & 0.613         & 0.710         & 0.698         & 0.925        & 0.017 & 0.051 & 97.9  & 91.4  & 93.3  & 97.9  & 95.1  \\
    \multirow{3}{*}{$\left(\hspace{-5pt}\begin{array}{l}
                \text{Whisper-L,}\\
                \text{TLTR-utt,}\\
                \text{7B-chat-LoRA}
                \end{array}\hspace{-5pt}\right)$}                &           & Sampling    & 0.594         & 0.691         & 0.679         & 0.920        & \B 0.020\tnote{\dag} & 0.069\tnote{\dag} & 97.8  & 88.7  & 91.7  & 98.6  & 94.2  \\ \cmidrule(l){2-14}
               & FCC       & Greedy    & 0.638\tnote{*}         & 0.726         & 0.714\tnote{*}         & 0.930\tnote{*}        & 0.016 & 0.047 & 98.3  & 92.2  & 94.2  & 98.5  & 95.8  \\ 
    &  ~(w/ predicted factors)         & Sampling    & 0.607         & 0.694         & 0.687         & 0.922        & \B 0.020\tnote{\ddag} & \B 0.070\tnote{\ddag} & 98.7  & 89.9  & \B 94.5  & 98.6  & 95.4  \\
    &           & GtS       & \B 0.640\tnote{+}         & \textbf{0.727}\tnote{+}         & \textbf{0.716}\tnote{+}         & 0.929\tnote{+}        & \B 0.020\tnote{\ddag} & 0.070\tnote{\ddag} & \B 99.0  & \B 93.2\tnote{+}  & \B 94.5\tnote{+}  & \B 99.2  & \B 96.5  \\ \cmidrule(l){2-14}
    & FCC       & Greedy    & 0.638\tnote{*}         & \B 0.727\tnote{*}         & \B 0.716\tnote{*}         & \B 0.931\tnote{*}        & 0.016 & 0.048 & 98.3  & 92.3  & 94.2  & 98.5  & 95.8  \\ 
    & ~(w/ golden factors)          & Sampling    & 0.607         & 0.693         & 0.687         & 0.922        & \B 0.020\tnote{\ddag} & \B 0.070\tnote{\ddag} & 98.7  & 89.9  & \B 94.5  & 98.6  & 95.4  \\
    &           & GtS       & \B 0.640\tnote{+}         & \textbf{0.727}\tnote{+}         & \textbf{0.716}\tnote{+}         & 0.929\tnote{+}        & \B 0.020\tnote{\ddag} & 0.067\tnote{\ddag} & \B 99.0  & \B 93.2\tnote{+}  & \B 94.5\tnote{+}  & \B 99.2  & \B 96.5  \\ \midrule
    \multicolumn{3}{@{}l}{Ground-truth captions}  & 1.000          & 1.000          & 1.000          & 1.000         & 0.020  & 0.071   & 99.1  & 99.1  & 99.6  & 99.4 & 99.3  \\  \bottomrule
  \end{tabular}
  \begin{tablenotes}
  \item[*/+] Statistically significant improvement from the model trained using the original captions with greedy search (*) / sampling (+).
  \item[\dag/\ddag] Statistically significant improvement from greedy search with the model trained using the original captions (\dag) / FCC (\ddag).
  \end{tablenotes}
  \end{threeparttable}
\end{table*}
The results on speaking-style captioning and style factor classification are listed in Table~\ref{tbl:main_results}.
First, as shown in the bottom line, it was found that the accuracies of factor class predictions from ground-truth captions using GPT were more than 99~\% in all the factors.
This suggests that GPT can extract speaking style factors from captions almost perfectly.
Compared to the captioning model, the first and the second rows, our model significantly outperformed the conventional model in all tasks\footnote{The StyleCap results were with sentence rephrasing augmentation, while ours were not. Our preliminary experiments found that rephrasing yielded no significant improvement in most setups.}.
As discussed in the following sections, all the components in our model were improved.
With the same model structures shown in the second, fourth, and sixth rows, FCC showed improvements in BLEU-4, ROUGE-L, METEOR, and BERTScore.
These differences were statistically significant according to the bootstrap-based test~($p$\textless .05, $n$=1000) except for FCC with predicted factors in ROUGE-L.
On the other hand, the performance of style factor classification was also improved but the differences were not significant.
These indicate that both the conventional captioning system and FCC can learn information of speaking-style factors from input speech, but FCC yields better captions.
One possible reason is that the factor prediction in the first step helps the model to focus on solving syntactic tasks in the caption generation step.
Comparing greedy and sampling-based search on caption and FCC-based models, the sampling search yielded improved output diversity as shown by distinct-1/2, but degraded both captioning and factor classification accuracies.
On the other hand, FCC with GtS showed no statistical degradation in either captioning or factor classification metrics from greedy search, while statistically improving caption diversity as shown by distinct-1/2~($p$\textless .05, $n$=1000).
These results show that the proposed FCC with GtS generates accurate and diverse captions.
Note that FCC with sampling-based search exhibited significant degradations on both tasks, which confirms that the sampling-based prediction of style factors is inappropriate.
It is finally worth mentioning that in all metrics, there is no statistical degradation between the proposed FCC with predicted factors and that with golden factors.

\setlength{\tabcolsep}{1.5mm} 
\begin{table}[t]
  \setlength{\tabcolsep}{1.4mm}
  \caption{Comparison of the speech encoders. The component used in StyleCap~\cite{StyleCap} is \underline{underlined}.}
  \label{tbl:enc_results}
  \vspace{-1em}
  \eightpt
  \centering
  \sisetup{detect-weight,mode=text}
  \begin{tabular}{@{}l*{2}{S[table-format=1.3]}*{4}{S[table-format=2.1]}@{}}
    \toprule
    & \multicolumn{2}{c}{Style captioning} & \multicolumn{4}{c}{Style-factor classification} \\ \cmidrule(lr){2-3}\cmidrule(l){4-7}
    & {B@4$\uparrow$} & {ROU$\uparrow$} & {Gender}   & {Pitch}   & {Speed}   & {Volume}  \\\midrule
    \underline{WavLM-B-plus}            & \underline{0.419}         & \underline{0.513}         & \underline{97.8}  & \underline{77.6}  & \underline{92.4}  & \underline{76.2} \\
    HuBERT-L                & 0.517         & 0.626         & 98.2  & 80.1  & 87.5  & \textbf{99.0} \\
    WavLM-L                 & 0.597         & 0.693         & \textbf{99.0}  & 88.4  & 92.5  & 97.8 \\
    Whisper-L               & \textbf{0.613}         & \textbf{0.710}         & 97.9  & \textbf{91.4}  & \textbf{93.3}  & 97.9 \\ \bottomrule
  \end{tabular}
\end{table}
\setlength{\tabcolsep}{1.5mm} 
\begin{table}[t]
  \caption{Comparison of the bridge networks.}
  \label{tbl:bridge_results}
  \vspace{-1em}
  \eightpt
  \centering
  \sisetup{detect-weight,mode=text}
  \begin{tabular}{@{}l*{2}{S[table-format=1.3]}*{4}{S[table-format=2.1]}@{}}
    \toprule
                                & \multicolumn{2}{c}{Style captioning} & \multicolumn{4}{c}{Style-factor classification} \\ \cmidrule(lr){2-3}\cmidrule(l){4-7}
                                & {B@4$\uparrow$} & {ROU$\uparrow$} & {Gender}   & {Pitch}   & {Speed}   & {Volume}  \\  \midrule
            W.sum\,+\,AvP         & 0.505         & 0.615         & 98.1  & 85.7  & 81.4  & 95.5  \\
            W.sum\,+\,CNN         & 0.383         & 0.501         & 98.1  & 74.3  & 79.4  & 88.6  \\
            \underline{W.sum\,+\,QF}          & \underline{0.561}         & \underline{0.668}         & \underline{\textbf{98.6}} & \underline{88.3}  & \underline{91.7}  & \underline{97.0}  \\
            TLTR-seg            & 0.578         & 0.666         & 97.8  & 88.4  & 89.9  & \textbf{98.2}  \\
            TLTR-utt            & \textbf{0.613}         & \textbf{0.710}         & 97.9  & \textbf{91.4}  & \textbf{93.3}  & 97.9  \\ \bottomrule
  \end{tabular}
\end{table}
\setlength{\tabcolsep}{1.5mm} 
\begin{table}[t]
  \setlength{\tabcolsep}{1.34mm}
  \caption{Comparison of the text decoders.}
  \label{tbl:dec_results}
  \vspace{-1em}
  \eightpt
  \centering
  \sisetup{detect-weight,mode=text}
  \begin{tabular}{@{}lc*{2}{S[table-format=1.3]}*{4}{S[table-format=2.1]}@{}}
    \toprule
                    &               & \multicolumn{2}{c}{Style captioning} & \multicolumn{4}{c}{Style-factor classification} \\ \cmidrule(lr){3-4}\cmidrule(l){5-8}
            Base    & LoRA       & {B@4$\uparrow$} & {ROU$\uparrow$} & {Gender}   & {Pitch}   & {Speed}   & {Volume}  \\  \midrule
            \underline{7B}      &               & \underline{0.578}         & \underline{0.684}         & \underline{97.7}  & \underline{89.0}  & \underline{\textbf{93.6}}  & \underline{\textbf{98.6}} \\
                    & \checkmark    & 0.602         & 0.695         & 98.1  & \textbf{91.5}  & 92.8  & 94.6 \\ \midrule
            7B-chat &               & 0.575         & 0.675         & \textbf{98.8}  & 88.3  & 92.9  & 96.5 \\
                    & \checkmark    & \textbf{0.613}         & \textbf{0.710}         & 97.9  & 91.4  & 93.3  & 97.9 \\ \bottomrule
  \end{tabular}
\end{table}

\subsubsection{Ablation study: Encoder, Bridge network, and Decoder}
To investigate the best components of the speaking-style captioning model, we compared several widely used modules in each component.
The base structure uses Whisper-L as the speech encoder, TLTR-utt as the bridge network, and 7B-chat-LoRA as the text decoder.
Each model was trained to output the original speaking-style captions and greedy search was used for decoding, which corresponds to the second row in Table~\ref{tbl:main_results}.
The training hyperparameters were the same as those in the main experiment.
Due to space limitations, we report here only BLEU-4, ROUGE-L, and four classification accuracies.

Comparisons of the speech encoders with WavLM-base-plus~\cite{WavLM}~(WavLM-B-plus), WavLM-large~(WavLM-L), and HuBERT-large~\cite{HuBERT}~(HuBERT-L) are detailed in Table~\ref{tbl:enc_results}.
WavLM-B-plus showed lower accuracy than the other encoders in pitch and volume prediction, same as the previous work~\cite{StyleCap}.
With regard to the remaining three encoders, Whisper-L achieved the best performance in both captioning and factor classification tasks.
This indicates that style captioning performance is highly dependent on the capability of the speech encoder, and that well trained large-scale models will provide better results.

Various bridge networks including the weighted sum of the intermediate layer outputs and average pooling~(W.sum\,+\,AvP), CNN~(W.sum\,+\,CNN), or BLSTM+Q-former~(W.sum\,+\,QF), and TLTR with segment-level outputs~(TLTR-seg, removing the temporal pooling as LTU-AS~\cite{LTU-AS}) are also compared in Table~\ref{tbl:bridge_results}.
The results indicate that capturing temporal context in the bridge is effective~(1, 2 vs. 3, 4, 5 rows), and that outputting a single-length embedding is more suitable than sequential embeddings in speaking style captioning and style factor classification~(4 vs. 5 rows).

Comparisons of the text decoders, LLaMA-2 7B or 7B-chat with and without LoRA, are shown in Table~\ref{tbl:dec_results}.
LoRA improved speaking-style captioning and pitch classification, but the performance difference between text decoders is smaller than that of the speech encoders and bridge networks.
This is because the variation in style captions was limited in this dataset, and so did not require a strong text generation module.

\section{Conclusion}
This paper presented FCC, a novel speaking-style captioning method.
FCC uses factor and caption terms that support the generation of captions conditioned by the predicted speech factors.
GtS, greedy-then-sampling, decoding initially evaluates the speech factors by maximum likelihood criteria and then generates captions, which yields diverse outputs while preventing factor prediction errors.
We also investigated the optimal Speech-Text model components.
Experiments showed that our method significantly outperformed the conventional style captioning model, and that FCC with GtS can generate accurate and diverse style captions while identifying style factors.

\bibliographystyle{IEEEtran}
\bibliography{output}

\end{document}